\documentclass[sigconf]{acmart}

\setcopyright{acmcopyright}
\copyrightyear{2019} 
\acmYear{2019} 
\acmConference[CIKM '19]{The 28th ACM International Conference on Information and Knowledge Management}{November 3--7, 2019}{Beijing, China}
\acmBooktitle{The 28th ACM International Conference on Information and Knowledge Management (CIKM '19), November 3--7, 2019, Beijing, China}
\acmPrice{15.00}
\acmDOI{10.475/123_4}
\acmISBN{123-4567-24-567/08/06}

\usepackage{enumitem}
\usepackage{bm}
\usepackage{xspace}
\usepackage{color}
\usepackage{mathtools}
\usepackage{subcaption}
\usepackage{booktabs}
\usepackage{multirow}
\usepackage{adjustbox}
\usepackage[ruled, vlined]{algorithm2e}
\usepackage[font=small,labelfont=bf,tableposition=top]{caption}
\usepackage{tikz}
\usepackage{balance}

\newcommand{\vpara}[1]{\vspace{0.08in}\noindent\textbf{#1}}

\newcommand{\eg}{{\sl e.g.\xspace}}
\newcommand{\ie}{{\sl i.e.\xspace}}

\newcommand{\etal}{{\sl et al.\xspace}}

\newcommand{\mc}{\mathcal}

\renewcommand{\phi}{\varphi}
\newcommand{\mbf}{\mathbf{f}}
\newcommand{\mbg}{\mathbf{g}}

\newcommand{\ctx}{\mc{C}}
\newcommand{\subc}[2]{{#1}^{(#2)}}
\newcommand{\subctx}[1]{{\ctx}^{(#1)}}

\newcommand{\rv}[1]{#1}
\newcommand{\rvn}[1]{#1}

\newcommand{\maple}{\mbox{\sf HyperMine}\xspace}

\newcommand{\lnkd}{\mbox{LinkedIn}\xspace}

\newcommand{\nop}[1]{}
\newcommand{\mquote}[1]{{``\emph{#1}''}}

\newcommand{\msr}[1]{\textbf{M#1}}
\newcommand{\smp}{\textit{Simplest}}
\newcommand{\grpby}[1]{\textit{Grp-by-#1}}
\newcommand{\clus}[1]{\textit{Clus-#1}}

\newcommand{\mlrr}{MLRR}
\newcommand{\marr}{MARR}

\newtheorem*{thm:eg*}{Example}
\newtheorem{thm:eg}{Example}

\begin{document}
\fancyhead{}

\title{Discovering Hypernymy in Text-Rich Heterogeneous Information Network by Exploiting Context Granularity}

\author{
Yu Shi$^{1\star}$, Jiaming Shen$^{1\star}$, Yuchen Li$^1$, Naijing Zhang$^1$, Xinwei He$^1$, Zhengzhi Lou$^1$, Qi Zhu$^1$, Matthew Walker$^2$, Myunghwan Kim$^3$, Jiawei Han$^1$
}
\affiliation{%
  \institution{$^1$Department of Computer Science, University of Illinois Urbana-Champaign}
}
\affiliation{%
  \institution{$^2$LinkedIn Corporation $\quad$ $^3$Mesh Korea}
}
\affiliation{%
  \institution{\{yushi2, js2, li215, nzhang31, xhe17, zlou4, qz3, hanj\}@illinois.edu }
}  
\affiliation{%
  \institution{
  $\quad$ $^2$mtwalker@linkedin.com $\quad$ $^3$mykim@cs.stanford.edu}
}





\begin{abstract}
Text-rich heterogeneous information networks (text-rich HINs) are ubiquitous in real-world applications.
Hypernymy, also known as \emph{is-a} relation or \emph{subclass-of} relation, lays in the core of many knowledge graphs and benefits many downstream applications. 
Existing methods of hypernymy discovery either leverage textual patterns to extract explicitly mentioned hypernym-hyponym pairs, or learn a distributional representation for each term of interest based its context. 
These approaches rely on statistical signals from the textual corpus, and their effectiveness would therefore be hindered when the signals from the corpus are not sufficient for all terms of interest.
In this work, we propose to discover hypernymy in text-rich HINs, which can introduce additional high-quality signals. 
We develop a new framework, named \maple, that exploits multi-granular contexts and combines signals from both text and network without human labeled data. 
\maple extends the definition of ``context'' to the scenario of text-rich HIN.
For example, we can define typed nodes and communities as contexts. 
These contexts encode signals of different granularities and we feed them into a hypernymy inference model.
\maple learns this model using weak supervision acquired based on high-precision textual patterns.
Extensive experiments on two large real-world datasets demonstrate the effectiveness of \maple and the utility of modeling context granularity. 
We further show a case study that a high-quality taxonomy can be generated solely based on the hypernymy discovered by \maple. 
\end{abstract}


\maketitle

{
\renewcommand{\thefootnote}{\fnsymbol{footnote}}
\footnotetext[1]{These authors contributed equally to this work.}
}


\section{Introduction}\label{sec::introduction}

\begin{figure}[!t]
	\centering\includegraphics[width=0.7\linewidth]{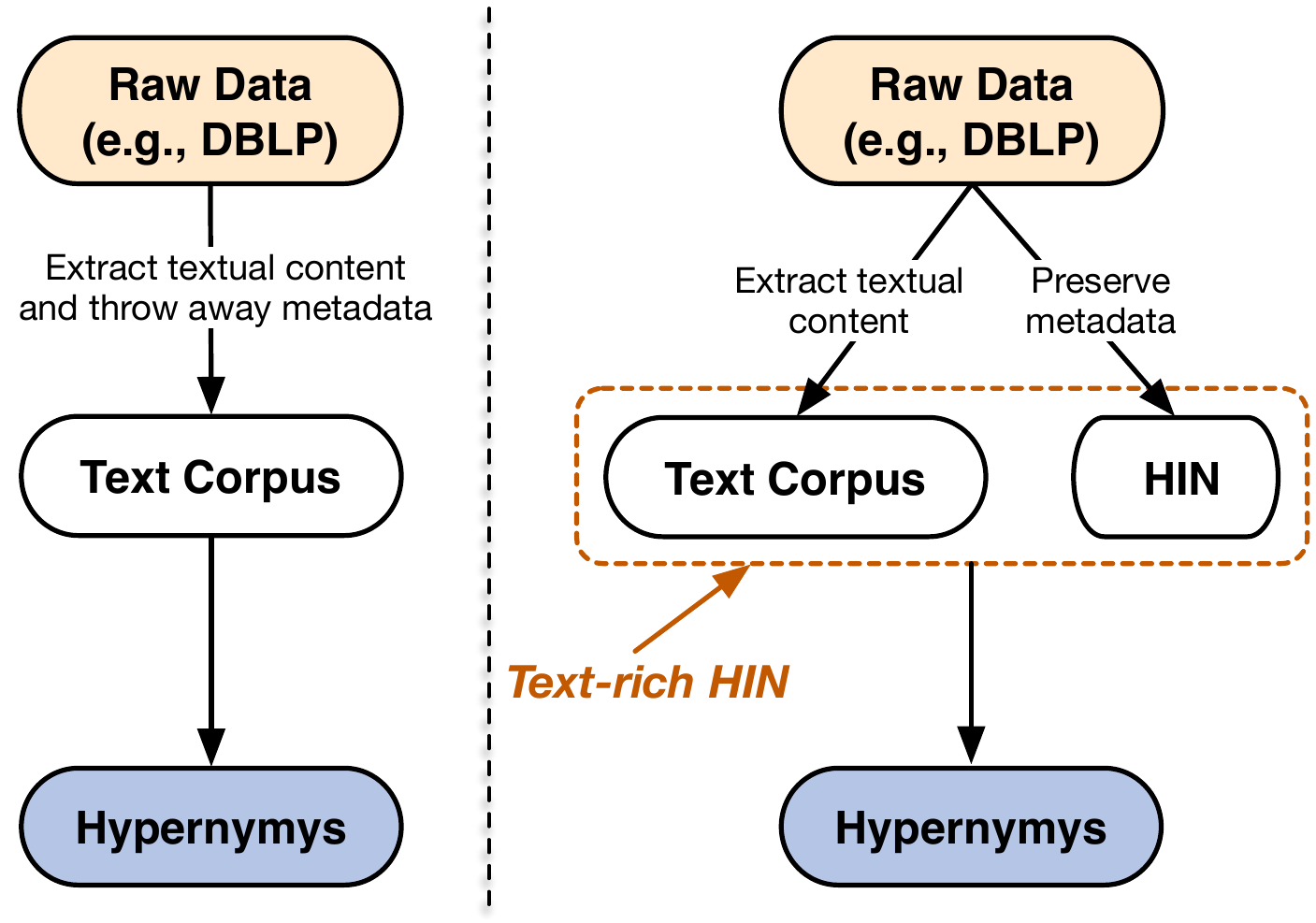}
	\caption[]{
		Comparison between previous work on hypernymy discovery only from text (left figure) and the proposed task of discovering hypernymy from text-rich HIN (right figure).
	}
	\vspace{-0.2cm}
	\label{fig:compare}
\end{figure}

Heterogeneous information network (HIN), as a powerful data model, has been widely studied since the past decade~\cite{sun2013mining, shi2017survey, yang2018meta}.
Many real-world HINs contain nodes associated with rich textual information~\cite{wang2015incorporating,  wang2017distant, yang2018similarity}, and we refer to them as \emph{text-rich HINs}. 
Typical examples of text-rich HINs include bibliographical networks (\eg, PubMed, DBLP) where nodes representing research papers are associated with their contents and social networks (\eg, Facebook, LinkedIn) in which nodes representing users are attached with their self-descriptive text. 
These text-rich HINs encapsulate both structured and unstructured information and empowers many downstream tasks such as document clustering~\cite{wang2015incorporating}, topic modeling~\cite{Shen2016ModelingTA}, and event detection~\cite{Zhang2016GeoBurstRL}.

Hypernymy, also known as \emph{is-a} relation or \emph{subclass-of} relation, is a semantic relation between two terms.
For example, \mquote{panda} is a \mquote{mammal} and \mquote{data structure} is a subclass of \mquote{computer science}.
Furthermore, we refer to a term $t_1$ as the \emph{hypernym} of term $t_2$ and $t_2$ as the \emph{hyponym} of $t_1$, if $t_2$ can be categorized under $t_1$. 
In the above case, \mquote{panda} and \mquote{data structure} are hyponyms while \mquote{mammal} and \mquote{computer science} are hypernyms.  
Hypernymy lays the foundations of many knowledge bases and knowledge graphs such as YAGO \cite{Suchanek2007YagoAC}, DBpedia \cite{Auer2007DBpediaAN}, and WikiData \cite{Vrandecic2014WikidataAF}.
Discovering high-quality hypernymy can also benefit many downstream applications such as question answering~\cite{yamane2016distributional, shwartz2016improving}, query understanding~\cite{Hua2017UnderstandST}, and taxonomy construction~\cite{mao2018end, Shen2018HiExpanTT, Zhang2018TaxoGenCT}.  

Existing work on hypernymy discovery focuses on detecting hypernymy pairs from massive text corpora. 
These methods typically fall into two categories --- pattern-based methods~\cite{hearst1992automatic, Shen2017SetExpanCS, Roller2018HearstPR} and distributional methods~\cite{weeds2004characterising, clarke2009context, shwartz2016improving}.
Pattern-based methods leverage high-precision textual pattern (\eg, ``X such as Y'') to extract hypernymy pairs. 
However, these patterns are usually language-dependent and have low recall, in the sense that they can only match explicitly stated hypernymy pairs in text. 
On the other hand, distributional methods, primarily based on the distributional inclusion hypothesis \cite{ZhitomirskyGeffet2005TheDI}, assume that the context of a hyponym is a subset of the context of a hypernym. 
A variety of textual contexts are defined, including nearby words in local window \cite{ZhitomirskyGeffet2005TheDI}, adjacent words in parse tree \cite{Baroni2010DistributionalMA}, or documents containing the term \cite{Shen2018AWS}, with different term-context weighting measures \cite{Sanderson1999DerivingCH, weeds2004characterising, lenci2012identifying}.
However, this approach usually performs poorly when only one type of textual context is used \cite{Shwartz2017HypernymsUS, Roller2018HearstPR}.
Moreover, to simultaneously model all types of contexts, this approach requires additional training hypernymy pairs which are often unavailable. 

\begin{figure}[!t]
\centering
  \begin{subfigure}[m]{0.485\linewidth}
    \centering\includegraphics[width=\linewidth]{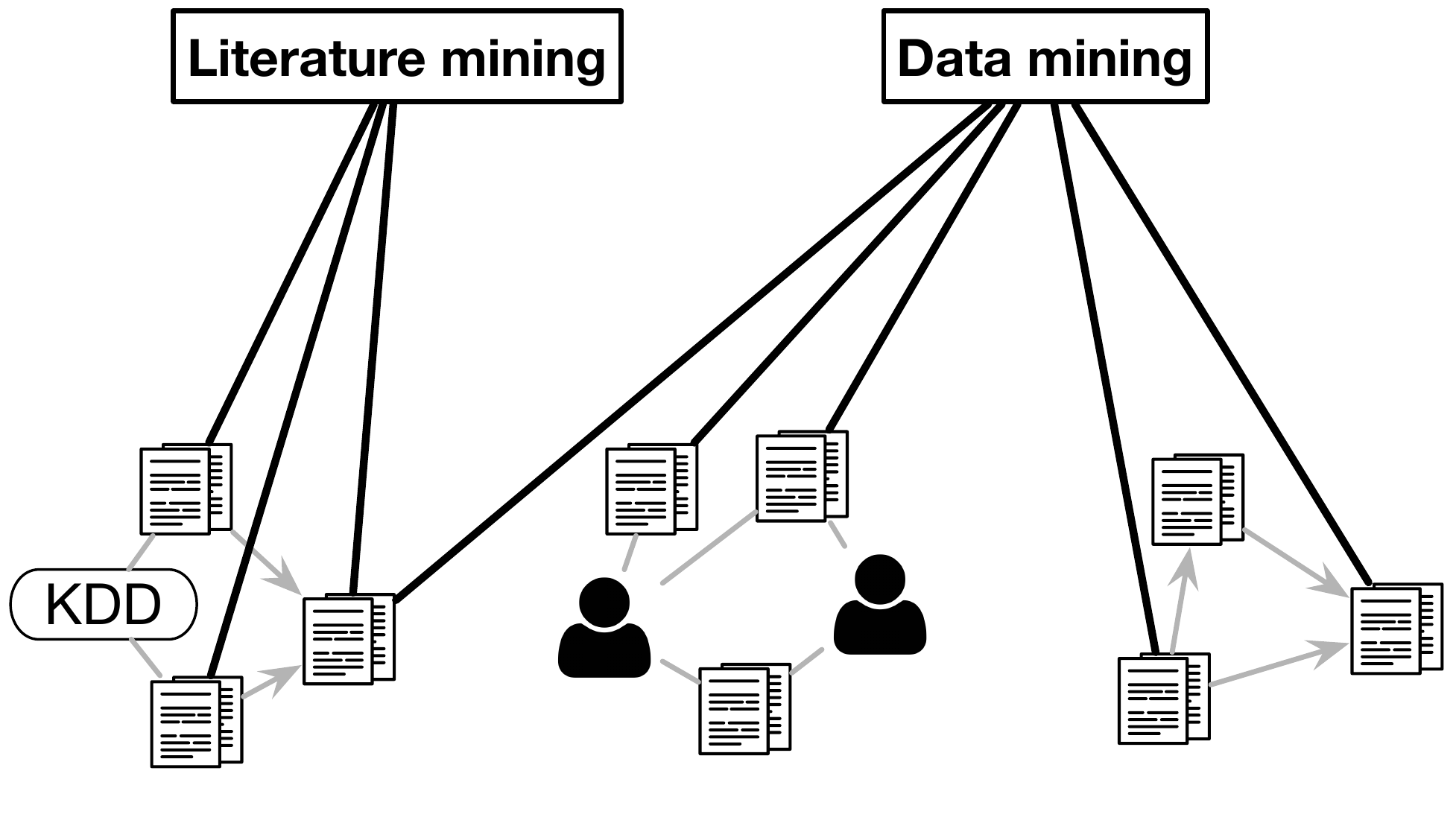}
    \caption{Each paper node is defined as a context unit.}\label{fig::intuition-smp}
    \vspace{-0.2cm}
  \end{subfigure}
  $\;$
  \begin{subfigure}[m]{0.485\linewidth}
    \centering\includegraphics[width=\linewidth]{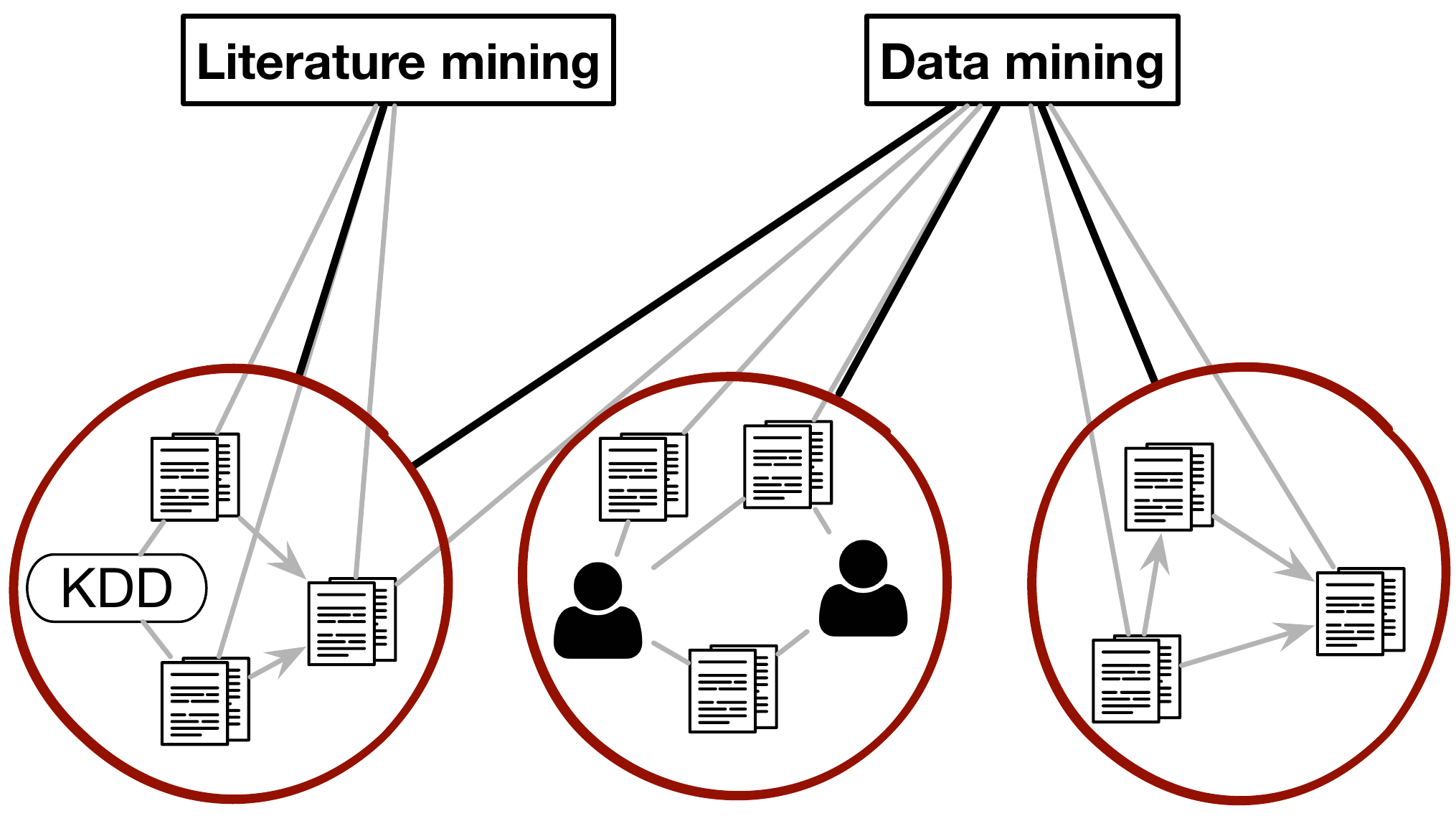}
    \caption{Each red circle is defined as a context unit.}\label{fig::intuition-coarse}
    \vspace{-0.2cm}
  \end{subfigure} 
 \caption[]{Example of the two definitions of context units with different granularities. By redefining a more coarse-grained context, we have the contexts of \mquote{Literature mining} to be a subset of the contexts of \mquote{Data mining}.}\label{fig::intuition}
\end{figure}

In this work, we propose to discover hypernymy from text-rich HINs.
The motivation for studying this problem is two-fold.
First, text-rich HIN is a better data model to preserve all the information in raw data (c.f. Figure~\ref{fig:compare}). 
For example, there are three existing competitions related to hypernymy discovery~\cite{Bordea2015SemEval2015T1, Bordea2016SemEval2016T1, CamachoCollados2018SemEval2018T9} and all of them use Wikipedia as their raw data. 
However, they only extract textual contexts from Wikipedia and simply discard all other structured information such as hyperlinks and Wikipedia categories, which causes severe information loss. 
Second, when the input corpus is small (\eg, the ACL Anthology where only NLP papers are included), existing methods may not work as there are less statistical signals such as co-occurrence and less chance of matching a textual pattern \cite{Yin2018TermDH}. 
However, if documents in this corpus are linked (\eg, by citation relations), we can model these documents using a text-rich HIN (\eg, the ACL Anthology Network \cite{Radev2009TheAA}).
Then, we can leverage the network part of the text-rich HIN to derive additional high-quality signals, which to some extent increases the applicability of hypernymy discovery method. 

\begin{figure}[!t]
  \centering
  \begin{subfigure}[m]{0.40\linewidth}
    \centering\includegraphics[width=\linewidth]{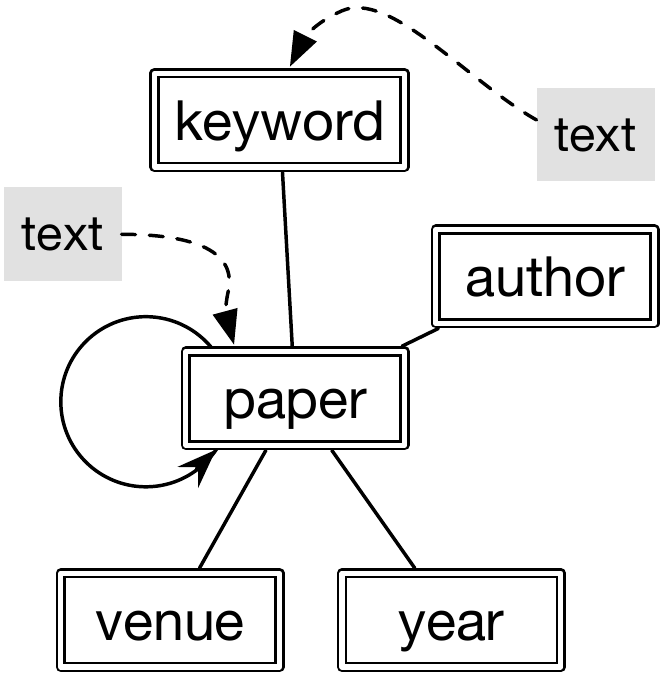}
    \caption{A bibliographical network (DBLP).}\label{fig::dblp-schema}
    \vspace{-0.2cm}
  \end{subfigure}
  $\;$
  \begin{subfigure}[m]{0.40\linewidth}
    \centering\includegraphics[width=\linewidth]{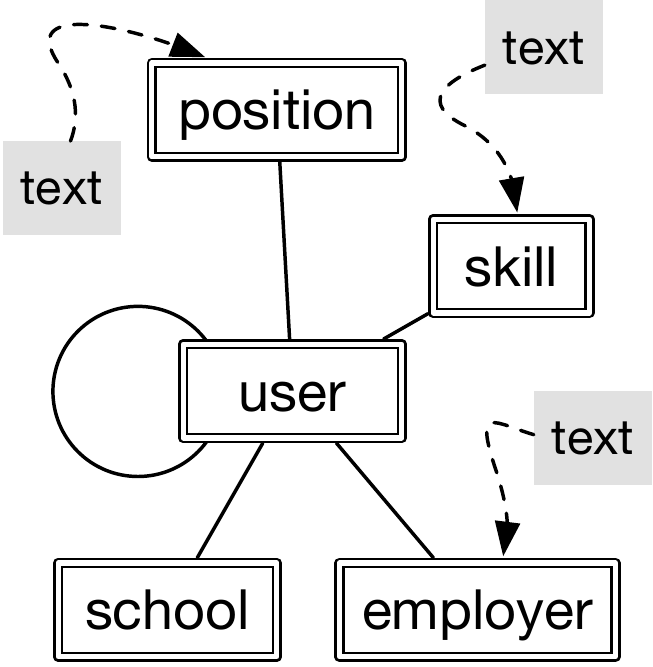}
    \caption{\rv{A professional social network (\lnkd)}.}\label{fig::lnkd-schema}
    \vspace{-0.2cm}
  \end{subfigure}
  \caption{The schemas of two text-rich HINs.}\label{fig::schema}
\end{figure}

To discover hypernymy from text-rich HINs, we develop a new unsupervised framework, named \maple, that exploits contexts of different granularities and combines signals from both text and network without any human labeled data. 
\maple extends the definition of ``context'' in distributional inclusion hypothesis (DIH) to include nodes in the network. 
For instance, in the DBLP network, we can define the context of each \textit{keyword} node to be the set of \textit{paper} nodes directly linked to it, as depicted in Figure~\ref{fig::dblp-schema}.
However, such a simple definition of context may not be proper for hypernymy discovery, as shown in the following example and Figure~\ref{fig::intuition-smp}:
\begin{thm:eg*}
We are given a pair of keywords ``literature mining'' ($t_1$) and ``data mining'' ($t_2$) and aim to predict whether it is a hypernymy pair based on DIH.
We denote all papers linked to ``literature mining''  as $\subctx{t_1}$ and those linked to ``data mining'' forms ${\subctx{t_2}}$.
If the DIH holds in this case, we should have ${\subctx{t_1}} \subseteq {\subctx{t_2}}$ as ``data mining''  is a hypernym of ``literature mining''.
However, a paper with ``literature mining'' linked as a keyword may not need to additionally tag the more general ``data mining'', and thus such scenario would violate the assumption of the DIH.
This can happen whenever the hypernym is too general, and the contextual units are too fine-grained.
\end{thm:eg*}

\maple resolves the above issue by redefining the ``context'' to be a group of semantically relevant papers, instead of each single paper.
Under this new definition, the desired property ${\subctx{t_1}} \subseteq {\subctx{t_2}}$ would hold as shown in Figure~\ref{fig::intuition-coarse}.
Such new contextual unit has more coarse granularity and is more general. 
In fact, we observe that the generality of a hypernymy pair (\ie, the overall generality of two terms involved in a hypernymy pair) is highly coupled with the granularity of the context, 
Therefore, we design \maple to incorporate multiple contexts with different granularities.

\begin{figure*}[!t]
 \centering\includegraphics[width=0.9\linewidth]{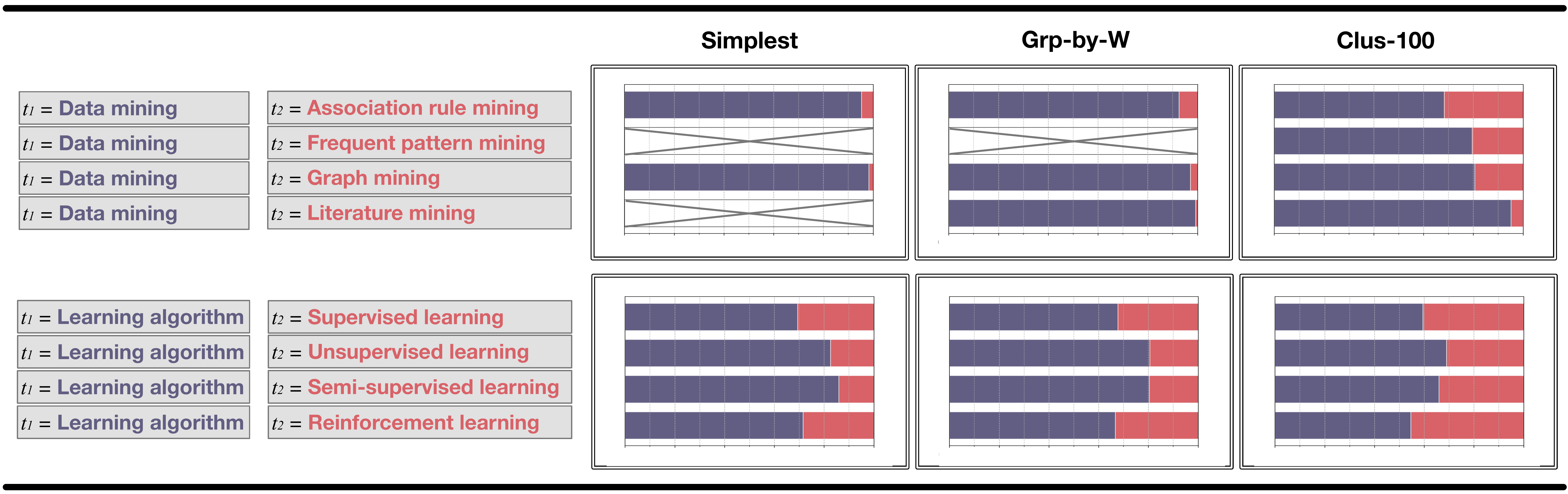}
 \caption[]{
Using the DBLP dataset, we demonstrate the observation that different hypernymy pairs should be discovered at different context granularities.
Each row in the plot corresponds to a hypernymy pair $t_1 \rightarrow t_2$, and each column corresponds to a context granularity.
In each bar, the length ratio between the left blue part and the right red part is proportional to the ratio between ${M_1}(t_1 \rightarrow t_2)$ and ${M_1}(t_2 \rightarrow t_1)$.
}\label{fig::case-study}
\end{figure*}

Finally, we need to combine signals from the textual part and the network part of the input data without resorting to human labeled training data.
To tackle this challenge, we first leverage textual patterns to extract a small set of quality hypernymy pairs from the textual part of text-HIN. 
The intuition is that while pattern-based methods have low recall, they tend to achieve high precision and thus will return a small set of quality hypernymy pairs. 
Then, we serve this small ``seed'' set into a hypernym inference model as weak supervision. 
This model takes two terms as input; calculates their nodewise features (\eg, embeddings in the text-HIN) and pairwise features (\eg, DIH measures based on multi-granular contexts), and returns a score indicating how likely they constitute a hypernymy pair. 
To better leverage information from the weakly-labeled set, we train the hypernymy inference model using a contrastive loss that penalizes the model whenever it predicts a higher score between a target term with its non-hypernym than the same target term with its true hypernym.

In summary, this paper makes the following contributions: 
(1) We propose to discover hypernymy from text-rich heterogeneous information networks (HINs), which introduces additional high-quality signals beyond text corpora; 
(2) We identify the impact of context granularity on distributional inclusion hypothesis (DIH) and propose the \maple framework that exploits multi-granular contexts and meanwhile combines signals from both textual and network data in text-HINs, and 
(3) We conduct extensive experiments to validate the utility of modeling context granularity and the effectiveness of leveraging HIN signals in hypernymy discovery. 
We further present a case study showing that \maple is able to discover high-quality hypernymy pairs for taxonomy construction.

\section{Problem Formulation}\label{sec::prob-def}

We first elaborate on some related concepts and useful notations and then formulate our problem.

\begin{definition}[\textbf{Heterogeneous Information Network}~\cite{sun2013mining}]
An \emph{information network} is a directed graph $G = (\mc{V}, \mc{E})$ with a node type mapping $\phi: \mc{V} \rightarrow \mc{T}$ and an edge type mapping $\psi: \mc{E} \rightarrow \mc{R}$. 
When $|\mc{T}| > 1$ or $|\mc{R}| > 1$, the network is called a \emph{heterogeneous information network} (HIN).
An HIN is referred to as a \textbf{text-rich HIN} if a portion of its nodes are associated with textual information that collectively constitute a \emph{corpus} $\mc{D}$.
\end{definition}

Given the typed essence of HINs, the network schema $\tilde{G} = (\mc{T}, \mc{R})$~\cite{sun2013mining} is used to abstract the meta-information regarding the node types and edge types in an HIN. 
Figure~\ref{fig::dblp-schema} illustrates the schema of the DBLP network, where a \textit{paper} node may have additional textual information from its content and a \textit{keyword} node may be associated with its description from Wikipedia.
Similarly, the schema of a social network in Figure~\ref{fig::lnkd-schema} consists of $5$ node types with \textit{skill}, \textit{employer}, and \textit{position} having textual information.

\begin{definition}[\textbf{Target Node Type and Vocabulary}]
A \emph{target node type} $T \in \mc{T}$ is a node type where each node of this type corresponds to a textual term (\ie, a word or a phrase)\footnote{Therefore, in the following, we use \mquote{term} and \mquote{node} interchangeably.}. 
We refer to the set of all such terms as the \emph{target vocabulary} $\mathcal{\Gamma}$.
\end{definition}

In the heterogeneous bibliographic network, DBLP, the target node type can be \textit{keyword}, and in a heterogeneous professional social network, LinkedIn, the target node type can be \textit{skill}.

\begin{definition}[\textbf{Context in DIH Measures}]
Measures based on distributional inclusion hypothesis are defined on a given domain of \emph{context} $\ctx$, over which each term in the target vocabulary $\mathcal{\Gamma}$ has a relevance distribution.
Given a term $t \in \mathcal{\Gamma}$ and a \emph{contextual unit} $c \in \ctx$, we denote the relevance between $t$ and $c$ as $r_c(t)$.
\end{definition}

Additionally, we denote the subdomain of the context that are relevant to term $t$ by $\subc{\ctx}{t} \coloneqq \{c \in \ctx \,|\, r_c(t) \neq 0\}$.
The primary intuition of measures based on distributional inclusion hypothesis is that $\subc{\ctx}{t_1}$ should include $\subc{\ctx}{t_2}$ if $t_1$ is a hypernym of $t_2$, and one widely-used DIH measure, $\mathit{WeedsPrec}$~\cite{weeds2004characterising}, is given by
\begin{equation}\label{eq::weedsprec}
\small
M_1(t_1 \rightarrow t_2) = \frac{\sum_{c \in \subc{\ctx}{t_1} \cap \subc{\ctx}{t_2}} r_c(t_2)}{\sum_{\subc{\ctx}{t_2}} r_c(t_2)}.
\end{equation}
The higher the $\mathit{WeedsPrec}$ score, the more likely $t_1$ is a hypernym of $t_2$.
In the traditional task of hypernymy discovery from a corpus, the typical definition of $C$ is the set of all words that co-occur with term $t$ in the corpus. 
Finally, we define our task as follows. 

\begin{definition}[\textbf{Problem Formulation}]
Given a text-rich HIN $G = (\mc{V}, \mc{E})$ and a target node type $T$ corresponding to a target vocabulary $\mathcal{\Gamma}$, the problem of \emph{hypernymy discovery from this text-rich HIN} aims to discover a list of hypernymy pairs with confidence scores where the hypernyms and the hyponyms are terms from $\mathcal{\Gamma}$.
\end{definition}


\section{Exploiting Context Granularity}\label{sec::observation}

\begin{figure*}[!t]
 \centering\includegraphics[width=0.98\linewidth]{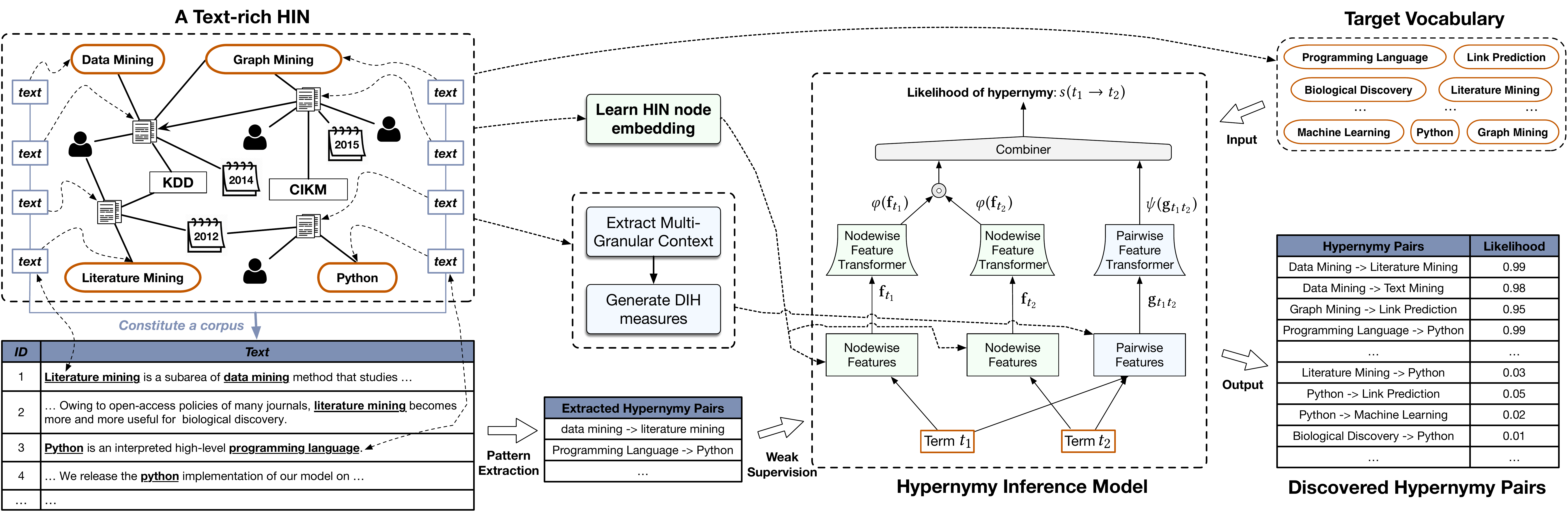}
 \caption[]{
The overview of the proposed \maple framework.
We discover hypernymy by exploiting rich signals from the network data besides the corpus.
A hypernymy inference model is trained using weak supervision extracted by high-precision patterns from the textual part of the text-rich HIN. 
A rich pool of features with comparably good term pair coverage is generated from the network part.
}\label{fig::overview}
\end{figure*}

In this section, we further illustrate our observation that different hypernymy pairs should be discovered with contexts of different granularities. 
Such observation motivates us to design the \maple framework in the next section. 
Figure~\ref{fig::case-study} presents the $\mathit{WeedsPrec}$ scores $M_1$, as calculated by Eq. (\ref{eq::weedsprec}), between eight hypernymy pairs computed at three contexts with different granularities --- \smp, \grpby{W}, and \clus{100}, where \smp~has the finest granularity and \clus{100} has the coarsest granularity. 
We will introduce their concrete definitions in Section~\ref{sec::node-feat}.
For each bar, the blue part on the left end is expected to be clearly longer than the red part on the right end, so that $\mathit{WeedsPrec}$ may be able to reveal $t_1$ as a hypernym of $t_2$ instead of the reversed order. 

In the first column (\smp{}) of the figure, the ratio between $M_1(t_1 \rightarrow t_2)$ and $M_1(t_2 \rightarrow t_1)$ cannot be visualized for two pairs involving ``data mining'' since their $M_1$ are trivially zero.
This undesirable result is the outcome of the fact that ``frequent pattern mining'' and ``data mining'' are never linked to the same contextual unit in the DBLP dataset to be described in Section~\ref{sec::data-dsc}, and likewise for ``literature mining'' and ``data mining''.
In the context represented by the second column (\grpby{W}),  $\mathit{WeedsPrec}$ is still trivially zero for one pair.
Fortunately, if we choose the context in the last column, $\mathit{WeedsPrec}$ can generate scores for all four pairs with hypernymy ``data mining''. 
As a result, by leveraging a coarser context, we can obtain more features and thus improve the recall.

However, the context in the last column (\clus{100}) is not always the best for all hypernymy pairs.
In the example of ``reinforcement learning'' ($t_1$) and ``learning algorithm'' ($t_2$), $M_1(t_1 \rightarrow t_2)$ and $M_1(t_2 \rightarrow t_1)$ are close to each other, which makes it hard to decisively assert ``reinforcement learning'' is a hyponym of ``learning algorithm''.
On the other hand, the distinction between $M_1(t_1 \rightarrow t_2)$ and $M_1(t_2 \rightarrow t_1)$ are much clear at the \smp{} context in the first column.
We interpret this result as the generality of a hypernymy pair is coupled with the granularity of the context, and hypernymy relations should, therefore, be revealed at multiple granularities.

Lastly, we emphasize that resolving the problem of the DIH shown in Figure~\ref{fig::intuition-smp} and Figure~\ref{fig::case-study} by exploiting context granularity is easier with the availability of HINs as input.
This is because, in an HIN, one can easily define semantically meaningful contextual units using explicit network structures such as grouping by a specific node type or more complex structures including meta-paths or motifs~\cite{shi2017survey, sun2013mining}.
Furthermore, one may also use network clustering methods to derive contextual units on a broad spectrum of granularities by varying the number of clusters.

\section{The \maple Framework}\label{sec::modeling}
We tackle the problem of discovering hypernymy from text-rich HINs by combining signals from both text and network using a hypernymy inference model.
This model infers the likelihood of a term $t_1 \in \Gamma$ being a hypernym of another term $t_2 \in \Gamma$. 
To learn this model without human labeled data, we leverage high-precision patterns to extract a set of hypernymy pairs from the corpus, which serves as the weak supervision. 
Then, we exploit the information encoded in HIN and generate a rich pool of multi-granular features for every single term and every term pair.
These features help to increase the recall. 
Particularly, we encode the network signals into nodewise features using HIN embedding and into pairwise features using the DIH-based measures under various context granularities.
Note that while the training of hypernymy inference model is weakly supervised, the whole \maple{} framework is unsupervised, as shown in Figure~\ref{fig::overview}.

\subsection{Weak Supervision Acquisition}
We generate weak supervision for the hypernymy inference model from the corpus $\mc{D}$ of the input text-rich HIN.
As a pioneering method, the Hearst pattern~\cite{hearst1992automatic} has been shown to have decent precision~\cite{wang2017short, wu2012probase, mao2018end}.
We use this method to extract a list $\mc{S} = \{(t_i^{\curlywedge}, t_i^{\curlyvee})|_{i=1}^{\mc{|S|}}\}$ of hypernymy pairs from the corpus $\mc{D}$.
In Section~\ref{sec::exp}, we will quantitatively evaluate the validity of this method for generating weak supervision.

\subsection{Nodewise Feature Generation}\label{sec::node-feat}
Network embedding has emerged as a powerful representation learning approach, which has been proven effective in many application scenarios~\cite{cui2018survey}.
A network embedding algorithm generally learns an embedding function $f: \mc{V} \rightarrow \mathbb{R}^{d}$ that maps a node to a $d$-dimensional vectorized representation.
In our \maple framework, we generate a feature $\mbf_v \coloneqq f(v)$ for each node $v \in \mc{V}$ using a network embedding algorithm designed for HINs from an existing study~\cite{shi2018easing}.
As each node of the target type in HIN represents a term, we can use this embedding function to derive the nodewise feature $\mbf_t$ for each term $t$ in the target vocabulary.

\subsection{Pairwise Feature Generation}
As introduced in Section~\ref{sec::introduction} and~\ref{sec::observation}, the DIH measures can be naturally extended to generate pairwise features from networks, whose power can be further unleashed by modeling context granularity.
For each term pair $(t_1, t_2) \in \Gamma \times \Gamma$, we generate a feature using one of the many DIH measures under one specific context definition.
Each DIH measure in our framework is henceforth referred to as a \emph{base DIH measure}.
In the following, we introduce the base DIH measures used in the \maple framework, together with various approaches to defining contexts with different granularities.

\subsubsection{Base DIH measures}
For a given context $\mc{C}$, we define the following four DIH measures.
\begin{itemize}[leftmargin=*, itemindent=8pt]
\item {\msr{1} ($\mathit{WeedsPrec}$~\cite{weeds2004characterising})} is one of the pioneering DIH measures defined as below:
$$\small M_1(t_1 \rightarrow t_2) = \frac{\sum_{c \in \subc{\ctx}{t_1} \cap \subc{\ctx}{t_2}} r_c(t_2)}{\sum_{\subc{\ctx}{t_2}} r_c(t_2)}.$$
\item {\msr{2} ($\mathit{invCL}$~\cite{lenci2012identifying})} is another widely used DIH measure which considers not only how likely $t_1$ is a hypernym of $t_2$ but also how unlikely $t_2$ is a hypernym of $t_1$. 
$$\small {M_2}(t_1 \rightarrow t_2) = \sqrt{\mathit{ClarkDE}(t_1 \rightarrow t_2) \cdot [1-\mathit{ClarkDE}(t_2 \rightarrow t_1)]},$$
$$\small \mathit{ClarkDE}(t_1 \rightarrow t_2) = \frac{\sum_{c \in \subc{\ctx}{t_1} \cap \subc{\ctx}{t_2}} \min(r_c(t_1), r_c(t_2))}{\sum_{c \in \subc{\ctx}{t_2}} r_c(t_2)}.$$
\item {\msr{3}} is a variant of \msr{2} and shares the same intuition as \msr{2}. 
$$\small {M_3}(t_1 \rightarrow t_2) = {\mathit{ClarkDE}(t_1 \rightarrow t_2) - \mathit{ClarkDE}(t_2 \rightarrow t_1)}.$$
\item {\msr{4}} is a symmetric distributional measure. Although it does not directly capture the inclusion intuition of the DIH, we use it to quantify the relevance of the term pair. 
$$\small {M_4}(t_1 \rightarrow t_2) = \frac{\sum_{c \in \subc{\ctx}{t_1} \cap \subc{\ctx}{t_2}} \min(r_c(t_1), r_c(t_2))}{|\ctx|}.$$
\end{itemize}

\subsubsection{Context Definition}
A simple way to define context given an HIN and a target node type is to let every node linked to nodes of the target type be a contextual unit.
We call the context $\mc{C}$ defined in this way the \smp{} context. 
As discussed in Section~\ref{sec::introduction}, atop the \smp{}, one may redefine contextual units by grouping the original ones that are semantically relevant.
With the availability of HIN data, we adopt the following two approaches to alternatively define the context in a broad spectrum of context granularities.

\begin{itemize}[leftmargin=*, itemindent=8pt]
\item \textbf{Define the context by explicit network structures.}
Many explicit network structures can be found in HINs such as the node types, the edge types, the meta-paths, and the meta-graphs~\cite{sun2013mining, shi2017survey}.
Using these structures, one can design methods to group the original contextual units in the \smp{} together to derive new contextual units.
In this paper, we adopt the most straightforward way and group together all contextual units linked to nodes of a specific node type.
We refer to this approach as \grpby{type} with \textit{type} being a specific node type.
As an example, in the DBLP dataset, a contextual unit in \grpby{author} is the collection of all papers written by a particular author.
We consider a term $t \in \mathcal{\Gamma}$ relevant to a contextual unit in \grpby{type} as long as $t$ is relevant to at least one original unit that is grouped into the new unit.
\item \textbf{Define the context by network clustering.}
Another way to derive semantically meaningful groups is by network clustering.
A great many clustering algorithms have been proposed for clustering HINs~\cite{sun2013mining, shi2017survey}.
With an intention to experiment with a simple algorithm while leveraging the rich information from HINs, we perform the classic $K$-means algorithm on the node features $\mbf_v \in \mathbb{R}^d$ to derive $K$ clusters.
Similarly,  a term $t \in \mathcal{\Gamma}$ is relevant to a cluster-based contextual unit as long as $t$ is relevant to at least one original unit in this cluster.
This approach is henceforth referred to as \clus{K}.
\end{itemize}

Given a term pair $(t_1, t_2) \in \mathcal{\Gamma} \times \mathcal{\Gamma}$, we compute a single score using each one of the base DIH measures together with one context type. 
Therefore, the pairwise feature $\mbg_{t_1 t_2}$ for term pair $(t_1, t_2)$ has the dimensionality equals to the number of base DIH measures times the number of context types. 
In this study, we focus our investigation on the benefit of introducing HIN signals and the utility of modeling context granularity, and we hence always set the relevance to be binary, \ie, $r_c(t) = 1$ if relevant and $0$, otherwise.

\begin{figure}[!t]
 \centering\includegraphics[width=.9\linewidth]{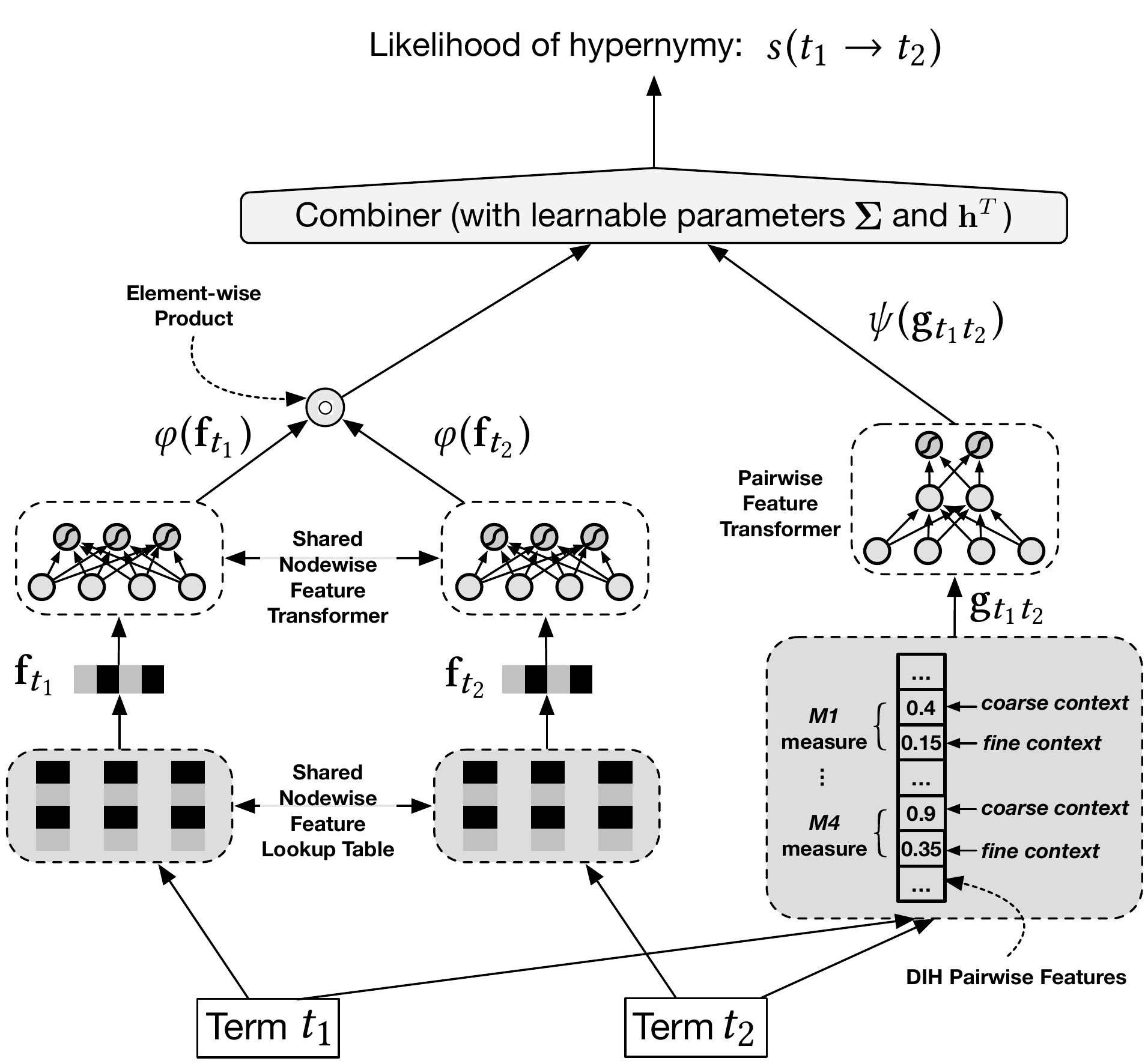}
 \caption[]{
Our proposed hypernymy inference model in \maple.
} \label{fig::nn-model}
\end{figure}

\subsection{Hypernymy Inference Model}\label{sec::nn-model}
We aim to obtain a model that calculates the likelihood of a pair of terms $(t_1, t_2)$ being a hypernymy pair, using weak supervision extracted from the text part and features from the network part of the input text-rich HIN.
The architecture of our hypernymy inference model has three major components, as depicted in Figure~\ref{fig::nn-model}.
The first component is a nodewise feature transformer $\phi(\cdot)$ that takes the raw nodewise feature $\mbf$ as input and transforms it into a new embedding space where the hypernymy semantics can be better captured.
We design this transformer to be a simple linear layer with dropout followed by a non-linear activation using $tanh(\cdot)$ function.
Following the core idea of the Siamese Network \cite{Bromley1993SignatureVU}, we apply the same nodewise feature transformer to both term $t_1$ and $t_2$. 
The second component is a pairwise feature transformer $\psi(\cdot)$ that acts upon the DIH-based pairwise features. 
Similarly, we design the pairwise feature transformer using a fully connected neural network with two hidden layers of size $N$ and $N/2$, where $N$ is the dimension of $\mathbf{g}_{t_1 t_2}$.
Again, we apply dropout for regularization and use $tanh(\cdot)$ function for activation. 
This pairwise feature transformer can capture the interaction across pairwise features derived from different contexts. 
The third component is a combiner that aggregates both nodewise and pairwise features after transformation and calculates the hypernymy score by
\begin{equation}\label{eq::hyper-score}
\small
s(t_1 \rightarrow t_2) = \phi(\mathbf{f}_{t_1})^{T} \bm{\Sigma}  \phi(\mathbf{f}_{t_2}) + \mathbf{h}^{T} \psi(\mathbf{g}_{t_1 t_2}),
\end{equation}
where $\bm{\Sigma}$ is a diagonal matrix and $\mathbf{h}$ is a vector.

To learn the parameters in both $\bm{\Sigma}$ and $\mathbf{h}$, we expect hypernymy pairs to have higher hypernymy scores than non-hypernymy pairs. 
Therefore, we use the following contrastive loss for model learning
\begin{equation}
\small
\mc{L} = \sum_{(t^{\curlywedge}, t^{\curlyvee}) \in \mathcal{{S}}} \sum_{t^{\times} \in \mc{N}(t^{\curlywedge})}  \max\left[0, 1- s(t^{\curlywedge} \rightarrow t^{\curlyvee}) + s(t^{\curlywedge} \rightarrow t^{\times})\right],
\end{equation}
where $\mc{N}(t^{\curlywedge})$ is a randomly sampled set of negative terms such that for any term $t^{\times}$ in the set, $(t^{\curlywedge}, t^{\times}) \notin \mc{S}$.
For each pair in $\mc{S}$, $L$ negative pairs are sampled.
This contrastive loss essentially penalizes the model whenever it predicts a higher score between a target term with its non-hyponym than the same target term with its true hyponym. 
By minimizing this loss, we can learn our hypernymy inference model.


\section{Experiments}\label{sec::exp}
In this section, we quantitatively evaluate the effectiveness of \maple on two real-world large text-rich HINs.
An additional case study is also presented using taxonomy construction as our downstream application.

\subsection{Data Description}\label{sec::data-dsc}
\smallskip
\noindent \textbf{Datasets.}
To the best of our knowledge, there is no standard benchmark dataset on hypernymy discovery in text-rich HIN. 
In this work, we use two large real-world HIN datasets\footnote{\scriptsize Available at \url{http://bit.ly/HyperMine-dataset}.} for the evaluation. 
\begin{itemize}[leftmargin=*, itemindent=8pt]
\item \textbf{DBLP} is a bibliographical network in the computer science domain, with five node types --- \textit{author} (A), \textit{paper} (P), \textit{keyword} (W), \textit{venue} (V), and \textit{year} (Y), and five edge types --- authorship, keyword usage, publishing venue, and publishing year of a paper, and the citation relation from a paper to another.
The text affiliated to a \textit{paper} node is the title of that paper, and the text associated with a \textit{keyword} node is the raw string of this keyword plus its Wikipedia page if the page exists.
We define \textit{keyword} to be our target node type. 
To generate a set of ground truth hypernymy pairs, we resort to the ACM Computing Classification System (CCS)\footnote{\scriptsize \url{https://www.acm.org/publications/class-2012}} which organizes computer science topics into a tree-structured taxonomy. 
A keyword in the vocabulary $\mathcal{\Gamma}$ is mapped to a topic term in CCS if they can be linked to the same Wikipedia entry using WikiLinker.
A positive hypernym-hyponym label is recorded if two keywords are mapped to two CCS terms that have ancestor-descendant relation in the CCS taxonomy.
Finally, we obtain $10,055$ positive hypernym-hyponym pairs.
Then, for each positive pair, we generate ten negative pairs by fixing the hypernym (hyponym) and randomly sampling five non-hyponym (non-hypernymy) keywords.
During such negative sampling process, a keyword is always randomly sampled from the set that can be mapped to CCS terms.

\item \textbf{\lnkd} is an internal profession social network that has five node types --- \textit{user}, \textit{skill}, \textit{employer}, \textit{school}, and \textit{position}, and five edge types -- users possessing skills, working for employers, attending schools, holding job positions, and being connected with other users.
\rv{The text associated to a \textit{skill}, a \textit{position}, and an \textit{employer} are respectively the Wikipedia page on this skill, users' descriptions for this position, and the job posting description created by this employer.}
The entire network is down-sampled to include only users from a major metropolitan area in the US as well as nodes and edges directly linked to these users.
\rvn{We further filter skills and keep the top $5,000$ regarding the number of users having a skill.}
We define \textit{skill} be the target node type.
Positive hypernym-hyponym pairs were curated by the company that \rv{owns} the data, and the label curation process is independent of our experiments.
We take the same process as described above to generate negative pairs.
\end{itemize}

The schemas of these two HINs are depicted in Figure~\ref{fig::schema}, and we summarize the \rv{statistics} of the datasets in Table~\ref{tab::data-stat}.

\begin{table}[t]
\centering
    \caption{Basic statistics of DBLP and LinkedIn datasets, where `M' represents million and corpus size $|\mc{D}|$ is the number of sentences.}\label{tab::data-stat}
    \vspace{-3pt}
\scalebox{.8}{
\begin{tabular}{| c || c | c | c | c | c | c |}
\hline
Dataset & $|\mc{V}|$ & $|\mc{E}|$ & $|\mc{T}|$ & $|\mc{R}|$  & $|\mathcal{\Gamma}|$ & $|\mc{D}|$ \\ \hline
DBLP   & 3,715,234 & 20,594,906 & 5 & 5  & 32,688 & 10,147,503 \\ \hline
LinkedIn  & M's & hundreds of M's & 5 & 5 & \rvn{5,000} & tens of M's \\  \hline
\end{tabular}
}
\end{table}

\begin{table*}[t]
\centering
    \caption{Quantitative evaluation results of hypernymy discovery from the DBLP and \rv{the \lnkd{} datasets}.}
    \vspace{-0.2cm}
    \label{tab::quant}
\scalebox{.85}{
\begin{tabular}{ l | c | c | c | c | c | c | c | c | c | c | c | c}
\toprule \hline
Dateset & \multicolumn{6}{c|}{DBLP}  &  \multicolumn{6}{c}{\lnkd} \\ \hline 
Metric & P@100 & P@1000 & Ma\marr & Mi\marr & Ma\mlrr & Mi\mlrr & P@100 &P@1000 & Ma\marr & Mi\marr & Ma\mlrr & Mi\mlrr \\ \hline \hline
Hearst~\cite{hearst1992automatic}  & 0.550 & 0.163 & 0.071 & 0.032 & 0.304 & 0.534 & 0.680 & 0.259 & 0.071 & 0.066 & 0.425 & 0.580 \\ \hline
LAKI~\cite{liu2016representing}    & 0.180 & 0.191 & 0.096 & 0.038 & 0.382 & 0.602 & 0.870 & 0.491 & 0.137 & 0.133 & 0.508 & 0.657 \\ \hline
Poincar\'e~\cite{nickel2017poincare} & 0.110 & 0.088 & 0.064 & 0.028 & 0.277 & 0.509 & 0.110 & 0.114 & 0.036 & 0.028 & 0.212 & 0.288 \\  \hline
LexNET~\cite{shwartz2016path}      & 0.580 & 0.337 & 0.121 & 0.044 & 0.463 & 0.542 & 0.660 & 0.529 & 0.129 & 0.098 & 0.534 & 0.605 \\ \hline \hline
\maple{}-wo-CG                     & 0.790 & 0.402 & 0.148 & 0.061 & 0.544 & 0.757 & \textbf{0.920} & \textbf{0.847} & 0.410 & 0.387 & 0.809 & 0.859 \\ \hline
\maple                             & \textbf{0.880} & \textbf{0.620} & \textbf{0.358} & \textbf{0.148} & \textbf{0.745} & \textbf{0.865} & {0.860} & {0.835} & \textbf{0.447} & \textbf{0.414} & \textbf{0.842} & \textbf{0.890} \\ \hline
\bottomrule
\end{tabular}
}
\end{table*}

\vspace{-6pt}

\subsection{Compared Methods}
We compare our framework with the following methods.
\begin{itemize}[leftmargin=*, itemindent=8pt]
\item \textbf{Hearst patterns} (\textbf{Hearst})~\cite{hearst1992automatic} is a classic pattern-based method for hypernymy discovery from text. 
\item \textbf{LAKI}~\cite{liu2016representing} is a document representation method which first learns a keyword hierarchy based on word embeddings (\ie, nodewise features) and DIH measures at the \smp{} context, and then assigns documents to this hierarchy. Since LAKI also incorporates both pairwise and nodewise features, we compare our framework with it in order to show that the higher expressive power of our hypernymy inference model indeed helps. 
\item \textbf{Poincar\'e Embedding} (\textbf{Poincar\'e})~\cite{nickel2017poincare} is an embedding learning algorithm in the hyperbolic space. It can embed an input taxonomy, represented as a directed acyclic graph (DAG), into a hyperbolic space, and then uses learned node embeddings to predict more hypernymy pairs in the DAG. We take it as a baseline because, to the best of our knowledge, this class of algorithms is the only existing ones that are relevant to hypernymy and take graphs or networks as input.
\item \textbf{LexNET}~\cite{Shwartz2016CogALexVST} is a state-of-the-art algorithm for hypernymy discovery from text. LexNet integrates dependency path based signals with distributional signals for predicting hypernymy. 
\item \textbf{\textsf{HyperMine-wo-CG}} is an ablated version of \maple which does not model context granularity and derives all the DIH measures based on raw context features.
\item \textbf{\textsf{HyperMine}} is the full version of our proposed framework\footnote{\scriptsize Code available at: \url{https://github.com/ysyushi/HyperMine}}.
\end{itemize}

\subsection{Evaluation Metrics and Experiment Setups}

\smallskip
\noindent \textbf{Evaluation Metrics.}
We report evaluation results using two precision metrics and four ranking metrics. 
The two precision metrics are precision at $k$ (\textbf{P@$k$)} with $k \in \{100, 1000\}$, which are computed as the number of positive pairs among the top-ranked $k$ pairs divided by $k$.
The four ranking metrics are macro mean average reciprocal rank (\textbf{Ma\marr}), micro mean average reciprocal rank (\textbf{Mi\marr}), macro mean largest reciprocal rank (\textbf{Ma\mlrr}), and micro mean largest reciprocal rank (\textbf{Mi\mlrr}).
To calculate these four ranking metrics, we first group all pairs sharing the same hypernym $t$ in the evaluation data.
Then, for each pair, its reciprocal rank (RR) is the reciprocal of the rank of this pair within the group.
Since there can be multiple positive pairs in a group, the average reciprocal rank for the group is the average over the RR's of all positive pairs, and the largest reciprocal rank is the largest RR among the RR's of all positive pairs.
Finally, we compute the macro mean and the micro mean across all groups to get the final four metrics, where the macro mean assigns uniform weights for all group when calculating the mean and the micro mean assigns weights proportional to the number of positive pairs in each group.
We also note that the optimal value the perfect model can achieve for Mi\marr{} and Ma\marr{} may be smaller than 1.
This can be explained with an example where a group has three positive pairs, then the highest average reciprocal rank for this group would be $(1/1+1/2+ \ldots + 1/6)/6 = 0.408 < 1$.
For all six metrics, greater value indicate better performance.

To obtain rank lists, we calculate a confidence score for each pair in the evaluation. 
All methods expect for baseline Hearst will directly return such confidence score.
For example, our hypernymy inference model will return a hypernymy likelihood $s(t_1 \rightarrow t_2)$ which can be naturally viewed as the confidence score.  
As for baseline Hearst, we simply assign a score of 1.0 for all of its extracted pairs and assign a score of 0.0 for all other pairs. 
When two pairs have the same scores, we randomly break the ties so that a model predicting all ties would get the same evaluation result as random guess.
For fairness across different runs of evaluation and across different models, we fix the random seed in the evaluation pipeline.

\smallskip 
\noindent \textbf{Experiment Setups.}
When determining context by explicit network structure, we use \grpby{A}, \grpby{V}, \grpby{W} for DBLP and \grpby{P}, \grpby{S}, \grpby{U} for \lnkd.
When determining context by clustering, we select two different values for $K$: 100 and 10000, which yields two contexts \clus{100} and \clus{10000}.
Therefore, we have totally 6 different contexts (5 derived contexts plus the simplest context), which, times 4 distinct base DIH measures, gives $4 \times 6 = 24$ pairwise features.

For both datasets, we learn the 128-dimension HIN node embedding using HEER \cite{Shi2018EasingEL}.
We tune all hyper-parameters of compared methods using 5-fold cross validation on our weak supervision dataset. 
For the hypernymy inference model in \maple, we use a neural network with one hidden layer of size 256 as the nodewise feature transformer. 
We set negative sampling ratio $L=10$, the dropout rate for $\phi(\cdot)$ to $0.7$, and dropout rate for $\psi(\cdot)$ to $0.1$.

\subsection{Quantitative Evaluation Results}\label{sec::quant-eval}

The main quantitative evaluation results are presented in Table~\ref{tab::quant}. 
Overall, the \maple{}-based methods outperform all baselines under all metrics in both datasets by large margins with only one exception for P@100 in the \lnkd{} dataset.
Furthermore, the full \maple{} model clearly outperforms \maple{}-wo-CG in DBLP and has a competitive performance with \maple{}-wo-CG in \lnkd{} dataset.

Notably, the state-of-the-art corpus-based method LexNET mostly excels among all baselines.
However, it still performs significantly inferior to \maple{} and \maple{}-wo-CG, which demonstrates the benefit of introducing network signals in the task of hypernymy discovery.
Also only taking corpus as the input, Hearst further underperforms LexNET on most metrics.
It is worth noting that the precision of Hearst drops drastically when the $k$ in P@$k$ changes from 100 to 1000 in both datasets.
In comparison, LexNET has P@100 similar to Hearst, while the former has clearly better P@1000.
This outcome further verifies the existing observation that Hearst tends to extract a limited number of term pairs, \ie, low recall, while the precision on the extracted pairs could be decent.

In addition, Poincar\'e is selected as a baseline because it represents a line of research that is both relevant to hypernymy and takes graphs or networks as input.
We find that Poincar\'e has the worst performance among all baselines, which might be because this algorithm is not designed for discovering hypernymy from data.
Furthermore, LAKI also generally performs worse than the other two \maple{}-based models with one exception for P@100 in \lnkd{}.
We analyze why it achieves better performance in \lnkd{} compared to in DBLP dataset in the next paragraph.

\begin{figure}[t]
  \centering
    \centering\includegraphics[width=\linewidth]{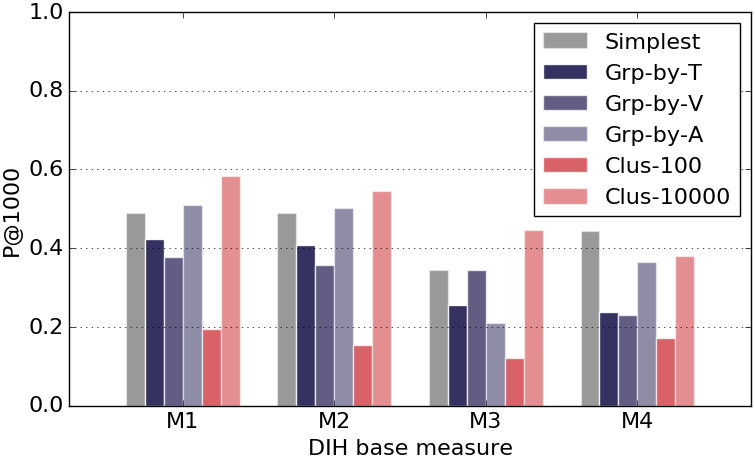}
  \caption{The importance of each DIH feature based on different base measures and different context granularities.}\label{fig::feat-imp}
\end{figure}

\begin{figure*}[t]
 \centering\includegraphics[width=.85\linewidth]{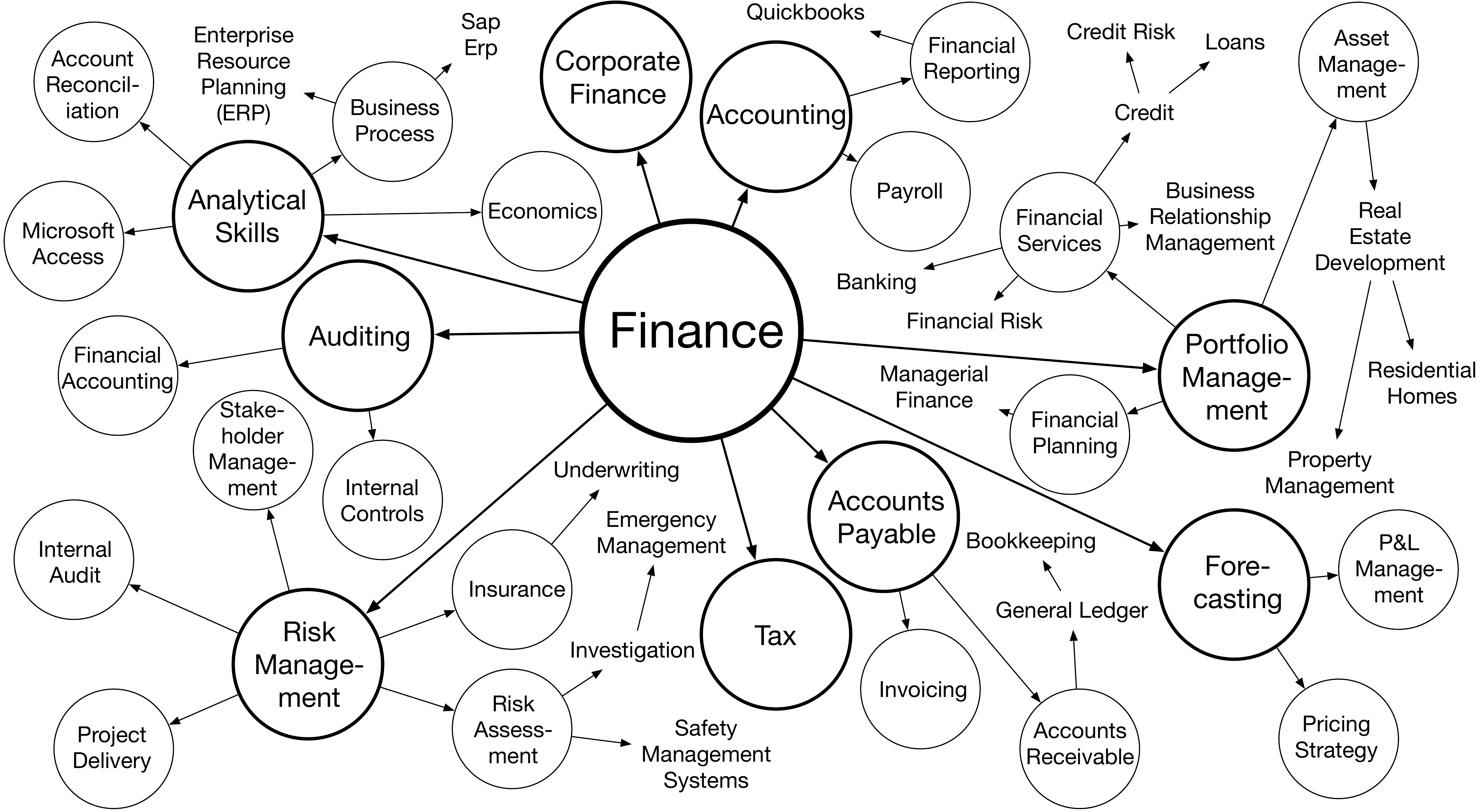}
 \caption[]{
A partial view of the skill taxonomy constructed from the hypernymy pairs discovered by \maple in \rv{the \lnkd{} dataset}.
 }\label{fig::vis-lnkd}
\end{figure*}

\smallskip 
\noindent \textbf{The same context granularity can have different importances in different datasets.}
\maple{} clearly outperforms its partial model \maple{}-wo-CG in DBLP dataset, which demonstrates the utility of leveraging multiple context granularities.
However, the comparison between \maple{} and \maple{}-wo-CG has mixed results in \lnkd~dataset.
We interpret this as the \smp{} context is too informative in \lnkd{} since each user is linked to more skills on average than each paper is linked to keywords.
As a result, introducing more features from other context granularities may not bring in a significant performance boost.
In fact, low-quality noisy features may even dampen the top-ranked pairs resulting in a lower precision, especially when $k$ is small.
This also explains the better performance of LAKI in \lnkd{}, since LAKI indeed uses a DIH measure at the \smp{} context.

\smallskip 
\noindent \textbf{Simultaneously leveraging pairwise features from multiple context granularities can introduce performance boost.}
In Figure~\ref{fig::feat-imp}, we plot out the evaluation result using each single network-based DIH feature on the DBLP dataset.
We only present the metric P@1000 and omit the other metrics due to space limitations, and that similar conclusion can be reached based on other metrics.
Comparing Figure~\ref{fig::feat-imp} with Table~\ref{tab::quant}, it can be seen that the proposed method using multi-granular context features achieves elevated performance compared with every single feature.
This finding corroborates our previous observation that different hypernymy pairs may be revealed from different context granularities.

\smallskip
\noindent \textbf{No context granularity is always the best even in the same dataset.}
From the performance of each single feature in Figure~\ref{fig::feat-imp}, it can be seen that not a context granularity is the best under all DIH base measures. 
For instance, \clus{10000} has the best performance when coupled with $M_1$, $M_2$, and $M_3$, while in the case of $M_4$, \smp{} is the best.
Also, while \smp{} is the best with $M_4$, it is even slightly worse than \grpby{A} when coupled with $M_1$ and $M_2$.

\subsection{Case Study: Taxonomy Construction}
In this section, we show that \maple can discover high-quality hypernymy pairs that are useful for taxonomy construction.
If we consider each discovered hypernymy pair as a directed edge, putting all pairs together will yield a graph potentially with cycles.
However, by definition, a taxonomy is restricted to be a directed acyclic graph (DAG).
Therefore, we resort to a simple heuristic algorithm used in many existing taxonomy construction studies~\cite{kozareva2010semi, velardi2013ontolearn}, which repeatedly finds one cycle in the current graph and then randomly breaks an edge.
We consider the resulting DAG as a crude taxonomy.

Due to the scalability limit of the above cycle breaking algorithm, we construct the initial graph with $500$ most popular skills and then keep $5000$ edges with the highest hypernymy scores.
We present the result after the cycle breaking algorithm in Figure~\ref{fig::vis-lnkd}, which includes the part of the DAG rooted by node \textit{Finance} and all of its hyponyms within the fourth-order neighborhood.

Since only $500$ top skills are left when constructing the graph, one should expect the recall of the taxonomy should be limited.
The recall aside, the constructed taxonomy has decent overall quality.
If we refer to a hyponym of $t$ that also has an edge from $t$ as a child of $t$, seven out of nine children of \textit{Finance} makes sense except \textit{Tax} and \textit{Analytical Skills}, where the latter two are related to \textit{Finance} but are not precisely its hyponyms.
One level deeper, descendants of the seven children are reasonable as well, which further corroborates the effectiveness of our proposed \maple{} framework.

As a final remark, even when such an unsupervised approach may not directly yield a perfect taxonomy, the discovered hypernymy pairs with confidence scores are still useful.
For example, when human labelers wish to expand an existing taxonomy to incorporate more nodes, they can seek recommendations from our \maple framework.


\section{Related Work}\label{sec::related-work}
In this section, we discuss related work for text-rich HINs and hypernymy discovery.

\vpara{Heterogeneous Information Network.} 
Heterogeneous information network (HIN) has been heavily studied for its ubiquity in real-world scenarios and its ability to encapsulate rich information~\cite{shi2017survey, sun2013mining, yang2018meta}.
Researchers have demonstrated that using HINs can benefit a wide range of tasks such as classification, clustering, recommendation, and outlier detection~\cite{shi2017survey, sun2013mining, zhuang2014mining}. 
Many real-world HINs are also text-rich with certain types of their nodes associated with additional textual information~\cite{yang2018similarity, deng2011probabilistic, wang2017distant}.
One typical example is the HIN with node type \textit{document} or \textit{research paper}. 
The content of each document or paper provides textual information highly relevant to this node, and such text-rich HINs have been studied in tasks such as clustering~\cite{wang2015incorporating}, topic modeling~\cite{Shen2016ModelingTA}, and literature search~\cite{Shen2018EntitySS}.

\vpara{Distributional Method for Hypernymy Discovery.}
Distributional methods constitute one major line of research for hypernymy discovery~\cite{wang2017short, turney2010frequency} and can be adapted to hypernymy discovery from network data.
Early studies proposed symmetric distributional measures for hypernymy discovery that only capture relevance between terms~\cite{lin1998information}. 
More recently, researchers have investigated into asymmetric measures based on the \textit{distributional inclusion hypothesis} (DIH) to comply with the asymmetrical nature of hypernymy~\cite{wang2017short, geffet2005distributional, zhitomirsky2009bootstrapping}.
Examples of popular DIH measures include WeedsPrec~\cite{weeds2004characterising}, APinc and balAPinc~\cite{kotlerman2010directional}, ClarkeDE~\cite{clarke2009context}, cosWeeds, invCL~\cite{lenci2012identifying}, and WeightedCosine~\cite{rei2014looking}.

Closely related to our effort in rectifying the DIH by modeling context granularity, several studies have also studied the validity of the DIH.
These studies have also suggested the DIH may not always hold accurate and proposed solutions orthogonal to ours~\cite{santus2014chasing, rimell2014distributional, roller2014inclusive}.
Santus et al.~\cite{santus2014chasing} propose an entropy-based measure SLQS that do not rely on the DIH, while some other studies suggested only certain units in the context should be used to generate features~\cite{rimell2014distributional, roller2014inclusive}.
We note that these approaches do not contradict with ours, because they are all based on the default context granularity, while we argue that DIH would hold at proper context granularities for each hypernym-hyponym pair.

\vpara{Pattern-based Method for Hypernymy Discovery.} 
Hearst, \etal~\cite{hearst1992automatic} pioneered the line of pattern-based hypernymy discovery methods which leverage hand-crafted lexico-syntactic patterns to extract explicitly mentioned hypernymy pairs from a text corpus.
A substantial number of methods have been proposed to extend the original six Hearst patterns~\cite{wu2012probase, etzioni2004web, kozareva2010semi}.
It has been shown that Hearst pattern based methods tend to achieve high precision with compromised recall~\cite{wang2017short, mao2018end, Roller2018HearstPR}.
Attempts have also been made to further improve the recall~\cite{anh2014taxonomy, snow2005learning, nakashole2012patty}.
In our framework, we use the straightforward Hearst pattern-based method to extract weak supervision pairs in the hope of yielding pairs with decent precision and without much additional engineering.

\vpara{Supervised Method for Hypernymy Discovery.}
With additional supervision available, researchers have proposed models to infer hypernymy based on the representation of a term pair~\cite{levy2015supervised, shwartz2016path, rei2018scoring}. 
Methods for deriving such representations include the aforementioned pattern-based methods and distributional methods as well as the compact, distributed representations generated from models such as word2vec~\cite{mikolov2013distributed}, GloVe~\cite{pennington2014glove}, and SensEmbed~\cite{iacobacci2015sensembed}.


\section{Conclusion and Future Work}
In this work, we propose to discover hypernymy from text-rich HINs, which avails us with additional rich signals from the network data besides corpus.
From real-world data, we identify the importance of modeling context granularity in distributional inclusion hypothesis (DIH).
We then propose the \maple framework that exploits multi-granular contexts and leverages both network and textual signals for the problem of hypernymy discovery.
Experiments and case study demonstrate the effectiveness of \maple as well as the utility of considering context granularity.

Future work can explore more methods to derive contextual units.
For example, we can use more complex structures (\eg, network motifs) and HIN-specific clustering methods to further unleash the utility of modeling context granularity.
Besides, it is also of interest to extend our framework to consider polysemy in datasets that possess such a characteristic.
Furthermore, in this work, we simply treat the textual part of a text-rich HIN as a corpus collection regardless of which node a particular piece of text is associated with.
Since a node and the text associated with it are likely to be relevant, we expect that building a more unified model by leveraging such signal could introduce additional performance boost to the task of hypernymy discovery.


\section*{Acknowledgements}\label{sec:ack}
This research is sponsored in part by U.S. Army Research Lab. under Cooperative Agreement No. W911NF-09-2-0053 (NSCTA), DARPA under Agreement No. W911NF-17-C-0099, National Science Foundation IIS 16-18481, IIS 17-04532, and IIS-17-41317, DTRA HDTRA11810026, grant 1U54GM114838 awarded by NIGMS through funds provided by the trans-NIH Big Data to Knowledge (BD2K) initiative, and LinkedIn Economic Graph Research Program.

\bibliographystyle{ACM-Reference-Format}
\bibliography{cited} 

\end{document}